%% file: main.tex
\documentclass[sigconf]{acmart}
\usepackage{CJKutf8}
\usepackage{algorithm}
\usepackage{algpseudocode}
\usepackage{graphicx,subfigure}
\usepackage{caption}
\usepackage{tabularx}
\usepackage{multirow}
\usepackage{float}
\usepackage{tabularx}
\usepackage{booktabs}
\usepackage{tabularx}
\usepackage{multirow}
\input{dat620-setup}

\begin{document}
\title{PFL-GAN: When Client Heterogeneity Meets Generative Models in Personalized Federated Learning}
\author{
    Achintha Wijesinghe$^1$,
    Songyang Zhang$^2$,
    Zhi Ding$^1$
}


\affiliation{$^1$University of California, Davis}
\affiliation{$^2$University of Louisiana at Lafayette}
\email{achwijesinghe@ucdavis.edu, songyang.zhang@louisiana.edu, zding@ucdavis.edu}

\begin{abstract}

Recent advances of generative learning models are
accompanied by the growing interest in
federated learning (FL) based on generative adversarial network (GAN) models.
In the context of FL, GAN can capture the underlying client data structure, and
regenerate samples resembling the original data distribution without compromising the private raw data. Although most existing GAN-based FL works focus on training a global model, 
Personalized FL (PFL) sometimes can be more effective in view of client data heterogeneity
in terms of distinct data sample distributions, feature spaces,
and labels. To cope with client heterogeneity in GAN-based FL, we propose a novel GAN sharing and aggregation strategy for PFL. The proposed PFL-GAN addresses the client heterogeneity in different scenarios. More specially, we first learn the similarity among clients and then
develop an weighted collaborative data aggregation. The empirical results through the rigorous experimentation on several well-known datasets demonstrate the effectiveness of PFL-GAN.

\end{abstract}


\maketitle


\section{Introduction}
Federated Learning (FL) has emerged as a promising direction in decentralized machine learning, offering robust privacy assurances without accessing raw data, while the generic machine learning implementations rely on centralized data access. In traditional FL frameworks~\cite{li2020federated}, a set of clients are connected to a central node (server), where each client can only access their local data. Usually, a global model is trained to satisfy most of the FL clients. Despite the successes, the classic FL approach still faces several fundamental challenges \cite{survey}: 1) poor performance to address client data heterogeneity; and 2) lack of solution personalization. As demonstrated in~\cite{FlFials1}, classical FL approaches, such as FedAvg~\cite{fedavg}, tend to underperform in the case of heterogeneous data, which is also known as nonindependent and nonidentically distributed (non-IID) data. This motivates the development of Personalized Federated Learning (PFL).

Personalized Federated Learning (PFL) is growing popular as a new branch of FL dedicated to client-specific data distribution and learning tasks. Different from classic FL, PFL focus on improving the performance of local models rather than training a global model, which shall have robust performances against data heterogeneity. For example, a meta-learning approach \cite{fallah2020personalized} is proposed to train a initial shared model which can be easily adapted to clients' local data. Beyond model sharing, the data augmentation can be alternatives to address data heterogeneity and solution personalization. Typical examples include FAug~\cite{FAug}, and FedCG~\cite{fedcgan}. In addition, clustering-based approaches can be another direction for PFL, where the relationships of clients are derived for grouping similar clients for better personalization. Among all the methods, generative learning models have also attracted significant attentions in handling data heterogeneity \cite{ours,our2}. Through learning local data distribution and regenerating synthetic samples for aggregation, FL based on generative adversarial networks (GAN) has shown promising performance in PFL. In \cite{byzantine}, a GAN-based framework is proposed to aggregate learning models and offers anomaly detection for PFL.

\begin{figure}[t]
\centering
\subfigure[]{\includegraphics[width=1.5in,height=0.3in]{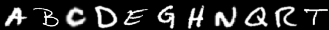}%
\label{fig3_1}}
\hfil
\subfigure[]{\includegraphics[width=1.5in,height=0.3in]{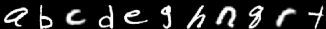}%
\label{fig3_2}}
\vspace{-3mm}
\caption{Example of Feature Heterogeneity with the Same Label: (a) capital handwriting letters, and (b) simple letters.}
\label{fig_capsim_int}
\end{figure}

Although existing PFL has already considered some client heterogeneity, most works focus on heterogeneous sample distribution, i.e., the so-called label-skewed ``non-IID" assumption, while failing to capture the realistic and pragmatic behavior encounters in realistic scenarios~\cite{rethinkheterogeneity}. For example, as shown in Fig. \ref{fig_capsim_int}, capital and simple handwriting letters show feature heterogeneity with label homogeneity. Some letters, such as `C', share similar features for capital/simple handwriting to benefit FL, while other letters like `B' may harm. In many other examples, such as music, movies, and books in commercial platforms, where different products may share feature correlation and some heterogeneity under similar labels, the recommendation system could benefit from evaluating data properties on more than one data types. Another way to consider this type of heterogeneity is to view other clients as byzantine compared to clients with the pivoted dataset. Therefore, we can expect the aforementioned methods to underperform when byzantine clients are present. How to uncover such data similarity and feature diversity jointly plays an important role in a more general PFL setup and heterogeneity consideration. Although clustering-based PFL may provide some hints to address these challenges, most of them assume symmetry and mutual neighbors while the realistic client's data distribution can be asymmetric. Considering the asymmetric Kullback-Leibler divergence (KLD) between two data distributions, two clients can be close with respect to one of the clients but not the other one. Therefore, it is worth envisioning the paramount importance of understanding the client's data distribution on the server instead of clustering clients in advance, which could benefit PFL regardless of the downstream tasks.

To address the client heterogeneity resulted from both sample distribution and feature space to personalize the FL, we investigate the generative learning models, and propose a novel \textbf{PFL-GAN} to provide a personalized solution for the clients. We summarize our contributions as follows:

\begin{itemize}
    \item We propose a novel GAN-based FL framework for a better-personalized solution for clients by securely understanding the client's private data distribution on the server side. Our proposed framework can work with heavy but practical data heterogeneity. 
    \item To the best of our knowledge, we are the first to address a more common real-world client data heterogeneity by selecting client representations from an amalgamation of datasets for the federation. 
    \item We propose a method to determine the similarity of client's private underlying data distributions before the model aggregation on the server, which helps secure the information from different client distributions, datasets, and byzantine clients. We decouple the need for original raw data to measure similarity between clients, unlike those clutering-based PFL in previous works.
    \item We present rigorous evaluations on several benchmarking datasets with different heterogeneous data setups against the state-of-the-art PFL methods, which demonstrate the effectiveness of PFL-GAN.
\end{itemize}

\section{Related Work}
Federated Learning ~\cite{fedavg} aims to enable a large amount of edge computing devices to jointly learn a model without data sharing. 
However, such classic FL techniques suffer from non-IID clients' data distribution ~\cite{FedBN}, which could result in considerable abate in classification accuracy. To address such challenges,
authors in~\cite{percent55} proposed global sharing of a subset of common data while the sharing of a subset of original raw data pose a severe privacy threat. To handle privacy,
synthetic data sharing is proposed in~\cite{syndata_1,syndata_2}. In addition to data sharing, an adaptive optimizer FedOpt is introduced for better convergence~\cite{Fedopt}.
Extended from FedAvg, FedProx is proposed in ~\cite{FedProx} to address the heterogeneity by introducing stability for FL by a proximal term. To handle the non-IID feature shift of clients, FedBN~\cite{FedBN} is proposed by local batch normalization. 
Another category of works applies GANs to address the non-IID data distribution, including ~\cite{cganfl,fedcgan,ours,fedgan_non1}.

Personalized FL is motivated by the poor convergence and the lack of personalization in highly non-IID data distribution~\cite{survey}. To develop PFL,
a GAN-based method is presented in RFA-RFD~\cite{byzantine} which uses knowledge distillation and anomaly detection for global model aggregation. Preferring the local performance to the global model, FedALA~\cite{fedala} proposes to capture the global model information locally by adaptive local aggregation. Despite the successes, the existing PFL approaches focus more on the heterogeneous sample distribution.
~\cite{rethinkheterogeneity}. In these approaches, different clients with different data sizes come under quantity skewness~\cite{quantskew,FedProx,quantskew2}. Other types of heterogeneity considered in existing works also include feature distribution skewness ~\cite{featureskew1,featureskew2} and label distribution skewness. As presented in
~\cite{extreme1}, in a pathological case where clients are having data from a single class, it prevents the global model convergence almost all the time. 
In realistic applications, however, data mixture with different feature spaces and different data types, sometimes can be more practical and challenging.
How to address such more general client heterogeneity remains an open question.

To provide a better solution for data heterogeneity, it is essential to understand and measure the data heterogeneity. Several recent studies, such as~\cite{hit1, FlFials1}, have proposed distance-based measurements, including the 1-Wasserstein distance and Earth Mover’s Distance~\cite{survey}. However, such measurements requires the evaluation on clients' private raw dataset, resulting in potential privacy leakage.
To overcome these issues, we propose to share GAN models with the server and design a novel strategy to measure the data similarity among clients without sharing raw data. 
Unlike existing prevailing Private Federated Learning (PFL) and Federated Learning (FL) techniques relying on model averaging, which is usually the non-optimal means of model aggregation~\cite{survey}, especially for non-IID data distributions~\cite{avgisnotthebest}, we leverage the proposed distance measurement to determine the number of samples needed to be generated from each GAN for aggregation, which enables a more efficient and effective process for data heterogeneity.

\begin{figure*}[htbp]
  \centering
\subfigure[Overall Structure of PFL-GAN]{ \includegraphics[width=1.2\columnwidth]{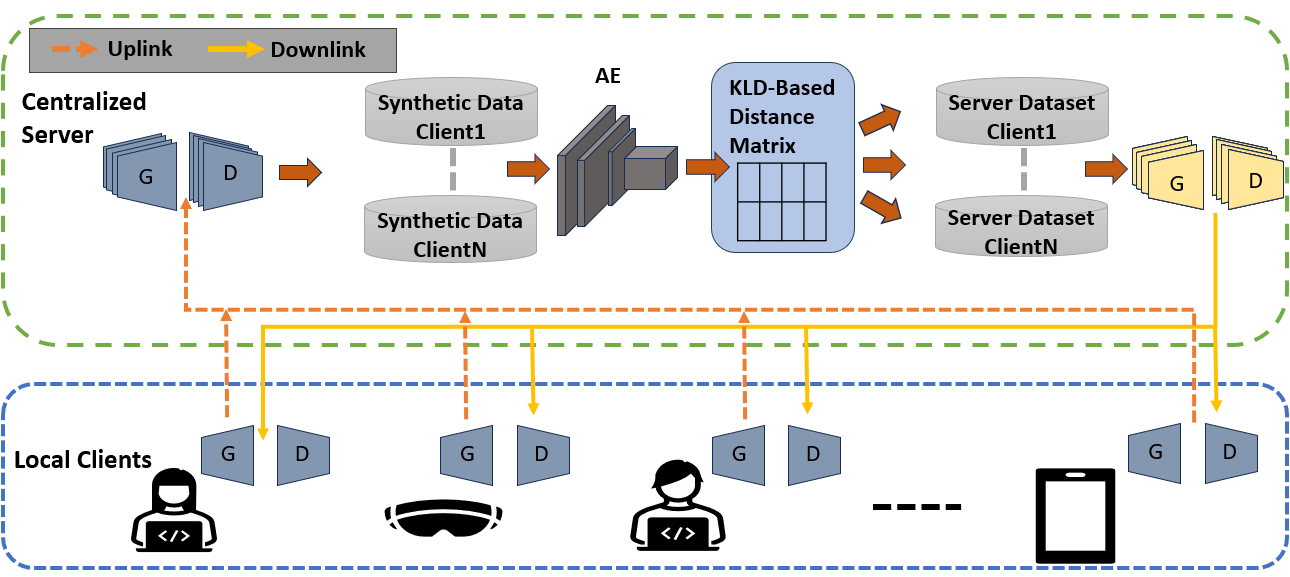}
\label{ova}}
\hfill
\subfigure[Structure of AE]{ \includegraphics[width=0.8\columnwidth]{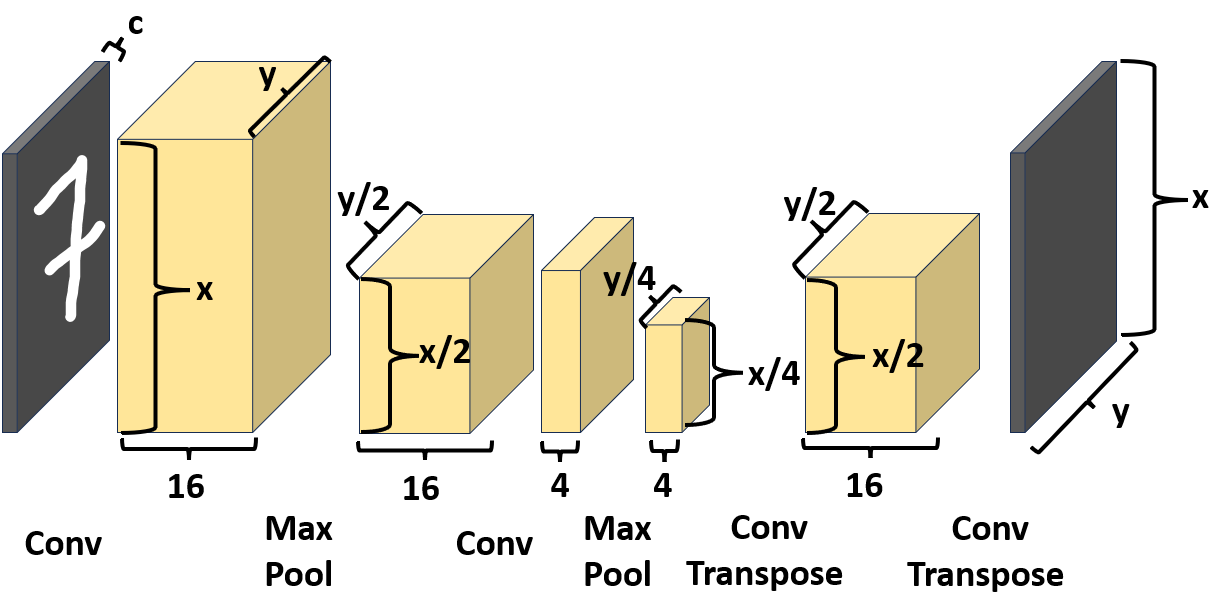}
\label{aes}}
  \caption{Overview of the proposed framework for personalized FL: On the client side, each client first initialize its cGAN model using local data and then share to the server. On the server side, each shared cGAN model first generate its synthetic IID dataset $I_{u_i}$. Then, a pre-trained AE is applied to learn the latent embedding for each $I_{u_i}$, based on which the similarity between clients are defined. We determine the number of synthetic data from other clients based on the similarities and aggregate to a new server dataset for each client $i$.
  Finally, we train a new cGAN per client according to their corresponding server dataset and share to the related local clients.}
  \vspace{-4mm}
  \label{overview}
\end{figure*}

\section{Method}
\subsection{Problem Statement} \label{prob}
In this work, our goal is to develop a novel PFL frameworks to deal with the client data heterogeneity. More specially, our proposed PFL-GAN shall be able to address various data heterogeneity, including label skewness, byzantine clients and different feature spaces.

 Suppose that the system contains $N$ clients and a centralized server. 
 The communication is only allowed between the server and each client, while the direct connection among clients are restricted due to data privacy. 
 For the evaluation purpose, we assume the image classification as the downstream task. Note that, our PFL-GAN shall be capable to any types of dataset. No global data is available at the server. Instead of training a global model, we are interested in personalize the model sharing and update for each client to achieve better local performance. To better illustrate the application background of our proposed PFL-GAN, we consider the following three scenarios as examples:

\begin{itemize}
    \item \textbf{Scenario-1}: All the $N$ clients are from the same dataset with the same feature and label space. For each client, we assign 300 samples for each class except 3 randomly selected classes, for which each selected class can only observe 15 samples. The Scenario-1 resembles the traditional non-IID label skewness.
    \item \textbf{Scenario-2}: In the second scenario,
    we consider $N$ clients and 2 totally different datasets (different feature and label spaces), such that $N/2$ clients are with dataset 1 and the remaining $N/2$ are with dataset 2. All $N$ clients have the same downstream tasks model architecture. For each client, there are 300 training samples for each class other than the 3 randomly-selected classes with 15 samples, similar to Scenario-1. The Scenario-2 resembles a more challenging non-IID distribution, which can be viewed to have byzantine clients in FL.
    \item \textbf{Scenario-3}: In Scenario-3, we consider that all the $N$ clients have the same label space but with two different feature spaces. For example, the two datasets can be the capital and simple handwriting letter dataset shown as Fig. \ref{fig_capsim_int}. We assign $N/2$ clients from each feature space. We follow the same data sample split as Scenario-1 and Scenario-2.
\end{itemize}
Note that, although we consider two different datasets and feature spaces in Scenario-2 and Scenario-3 respectively, our PFL-GAN shall be easily extended to the scenarios with more than two datasets/spaces. We shall present the results for all these scenarios later in Section \ref{experiment}.

\subsection{PFL-GAN}
In this section, we introduce PFL-GAN, which is designed to address the practical data heterogeneity among clients.

\subsubsection{ Overall Structure of PFL-GAN}
The overall framework of PFL-GAN is illustrated as Fig. \ref{ova}. To initialize the PFL-GAN training, each client $i$ first train a local conditional GAN (cGAN), i.e., the generator $G_{u_i}$ and discriminator $D_{u_i}$, based their raw data. Then, each initial GAN model is transmitted to the server for aggregation. On the server side, an IID synthetic data $I_{u_i}$ is generated based on $G_{u_i}$. Then, we define a similarity measurements among each $G_{u_i}$ with an auto-encoder (AE), and determine the ratio of synthetic data that shall be used for aggregation to generate a new server dataset $T_{u_i}$ for client $i$. The structure of AE and the definition of similarity measruements will be discussed later in Section 3.2.2.
If the property of $G_{u_j}$ is closer to $G_{u_i}$, the refined dataset $T_{u_i}$ contains more samples from $I_{u_i}$. Finally, based on each refined server dataset $T_{u_i}$, a new cGAN model, i.e., $G_{g_i}$ and discriminator $D_{g_i}$, can be trained and then shared to client $i$ to generate synthetic data for downstream tasks. Since each refined cGAN favor the synthetic distribution from similar clients, it is personalized for each client for their own local data distribution and tasks. 

\subsubsection{ Auto-Encoder Guided Framework for Client Similarity}
We now provide more details about how we calculate the client similarities and determine the ratio of synthetic samples each $I_{u_j}$ should contribute to client $i$.

Suppose that we have an AE with an encoder ($\mathcal{E}$) and a decoder ($\mathcal{D}$) pre-trained for image reconstruction with the structure shown as Fig. \ref{aes}. The AE is trained with a single dataset (e.g. MNIST) by minimizing the error between input and reconstructed images. Figure~\ref{aes} illustrates the AE architecture. We implement AE using convolutional layers with a latent layer of 256 ( after flattening) for our experiments. Therefore, the latent describes 256 features of a given dataset. 

Let $\mathcal{X}_i$ denote the private data distribution for client $i$, $i=1,\cdots, N$. The pre-trained $\mathcal{E}$ is applied to map each training sample to a low-dimensional latent representation $l$, i.e., for any $x_j \in \mathcal{X}_i$
\begin{equation}
    \mathcal{E}(x_j) \mapsto l_j.
\end{equation}

Then, we convert the latent embeddings $l$ to probabilities using the $Softmax$ function, i.e.,
\begin{equation}
    p_j = Softmax(l_j).
\end{equation}
Here, $p_j$ represents the probability vector corresponding to the probabilities of each feature in the feature space spanned by the $\mathcal{E}$'s latent space for the corresponding image. 

If $l^{1 \times R}$, the probability distribution of the client $i$ distribution with respect to the feature space $l$ is taken by
\begin{equation}
\label{eq:3}
    P_i = \frac{\sum_{x \in \mathcal{X}_i}\mathcal{E}(x)}{\sum_{k=1}^{R}\sum_{x \in \mathcal{X}_i}\mathcal{E}(x)}.
\end{equation}

Then the distance between client $i$ and client $j$ denoted by $d_{ij}$ is calculated with respect to the Kullback–Leibler divergence (KLD) as follows:
\begin{equation}\label{kld}
    d_{ij} = KLD(P_i||P_j).
\end{equation}

With the client similarity, we generate a new dataset $T_{u_i}$ for each client $i$ to train a personalized cGAN. We use $d_{ij}$ to decide the number of samples that needs to generate from client $j$ to support client $i$. More specially, we consider the truncated radial basis function (RBF) kernel \cite{10180031,zhang2022multilayer} to define the this ratio of samples that $T_{u_i}$ should contain from $I_{u_j}$, i.e.,
\begin{equation}
\label{eq:7}
  S_{ij} =
    \begin{cases}
      c\exp({-\frac{d_{ij}^2}{2\sigma_i^2})} & \text{if $d_{ij} < \tau_i$ }\\
      0 & \text{otherwise}
    \end{cases}       
\end{equation}
where $c$ is a predefined constant, the threshold $\tau_i$ is calculated by
\begin{equation}
\label{eqn:tau}
    \tau_i = \frac{\sum_{j=1}^{N}d_{ij}}{(N-1)}
\end{equation}
and $\sigma^2$ is computed by
\begin{equation}
    \sigma_i^2 = \frac{\sum_{j=1}^{N}(d_{ij}-\tau_i)^2}{(N-1)}.
\end{equation}

Note that, $\tau$ groups the clients' synthetic datasets into two clusters: one for aggregation and the other for ignorance. Here, we select the $\tau$ based on the mean distances as a basic guideline. More advanced approaches can be applied to select a more efficient $\tau$. More details will be discussed in Section \ref{taus}.

By generating a synthetic IID server dataset by aggregating from different $G_{u_i}$, we hide each client's true sample distribution. Meanwhile, the synthetic server dataset capture the overall feature space more accurately integrated with synthetic samples from similar clients. 

Finally, we generate $\lfloor S_{ij}\rfloor$, i.e., the number of samples, from client $j$'s synthetic dataset $I_{u_j}$ to create a new server dataset $T_{u_i}$ for client $i$ to train a personalized cGAN $G_{g_i}$ according to the Eq. (\ref{eq:7}). We then share the corresponding GAN models $G_{g_i}$ with the corresponding client $i$, after which the client uses the received GAN to generate new training samples and aggregate them with the available private training data to train the downstream model locally. The PFL-GAN detailes in presented in Algorithm~\ref{alg:1}.

\begin{algorithm}[tb]
\caption{PFL-GAN training algorithm}
\label{alg:1}
\begin{algorithmic}[1] 
\For {each client $i$ on the client-side}
    \State Initialize generator $G_{u_i}$ and discriminator $D_{u_i}$ using original local dataset
\EndFor



\State Share $G_{u_i}$ and $D_{u_i}$ with the server.
\For {each client $i$ on the server-side}
    \State Generate an IID dataset: $I_{u_i}$ using $G_{u_i}$ 
\EndFor
\For {each dataset $I_i$ }
    \State Calculate $P_i$ using the Eq. (\ref{eq:3})

\EndFor
\For {each client $i$ on the server-side}
    \State Calculate $d_{ij}$ based on Eq. (\ref{kld})
    \State Calculate $S_{ij}$ based on Eq. (\ref{eq:7})
    \State Generate a dataset $T_{u_i}$ using $S_{ij}$ and $d_{ij}$
    \State Train a generator $G_{g_i}$ and discriminator $D_{g_i}$ using $T_{u_i}$
    \State Share the $G_{g_i}$ with the client $i$
\EndFor
\For {each client $i$ on the client-side}
    \State Train $Cl_{u_i}$ (Downstream Classifier) using $G_{g_i}$'s generated data and original private data
\EndFor

\end{algorithmic}
\end{algorithm}

\section{Experiments} \label{experiment}
In this section, we present the experimental results to evaluate the effectiveness of our proposed method. 
\subsection{Experiment Setup}
We evaluate the proposed method against existing FL frameworks, including FedAvg~\cite{fedavg}, FedProx~\cite{FedProx}, FedOptim~\cite{Fedopt}, RFA-RFD~\cite{byzantine}, and FedBN~\cite{FedBN} in several benchmarking image datasets, such as MNIST~\cite{deng2012mnist}, FMNIST~\cite{fashionMnist}, and EMNIST~\cite{emnist}.

As aforementioned in Section \ref{prob}, we consider three scenarios and data splitting to evaluate the performance of PFL-GAN with different types of data heterogeneity:
\begin{itemize}
    \item For Scenario-1 (Label Skewness), we set $N=20$ clients and test in the MNIST dataset.
Each client's private dataset consists of all 10 MNIST classes with 300 samples per class except for 3 randomly selected classes. These three random classes contain 15 samples per class. 
\item For Scenario-2 (Byzantine), we set $N=20$ where 10 random clients represent the MNIST dataset and the rest of the 10 clients represent the FMNIST dataset. The sample distribution for each client is the same as Scenario-1.
\item For Scenario-3 (Heterogeneous Feature Space), we consider 10 capital letters and 10 simple letters of the corresponding capital letters in the EMNIST dataset. We randomly set $N=10$ with 5 capital letter clients and 5 simple letter clients. Each client has the same sample distribution per class as Scenario-1 and Scenario-2.
\end{itemize} 

We implemented the training and test via torch 1.13 on a server with NVIDIA TITAN V GPUs with an Intel Xeon Silver 4110 CPU (32 cores) with a memory size of 64 GB.

\subsection{Results of Scenario-1}
\begin{figure}[t]
\centering
\includegraphics[width=0.4\textwidth, height=0.3\textwidth]{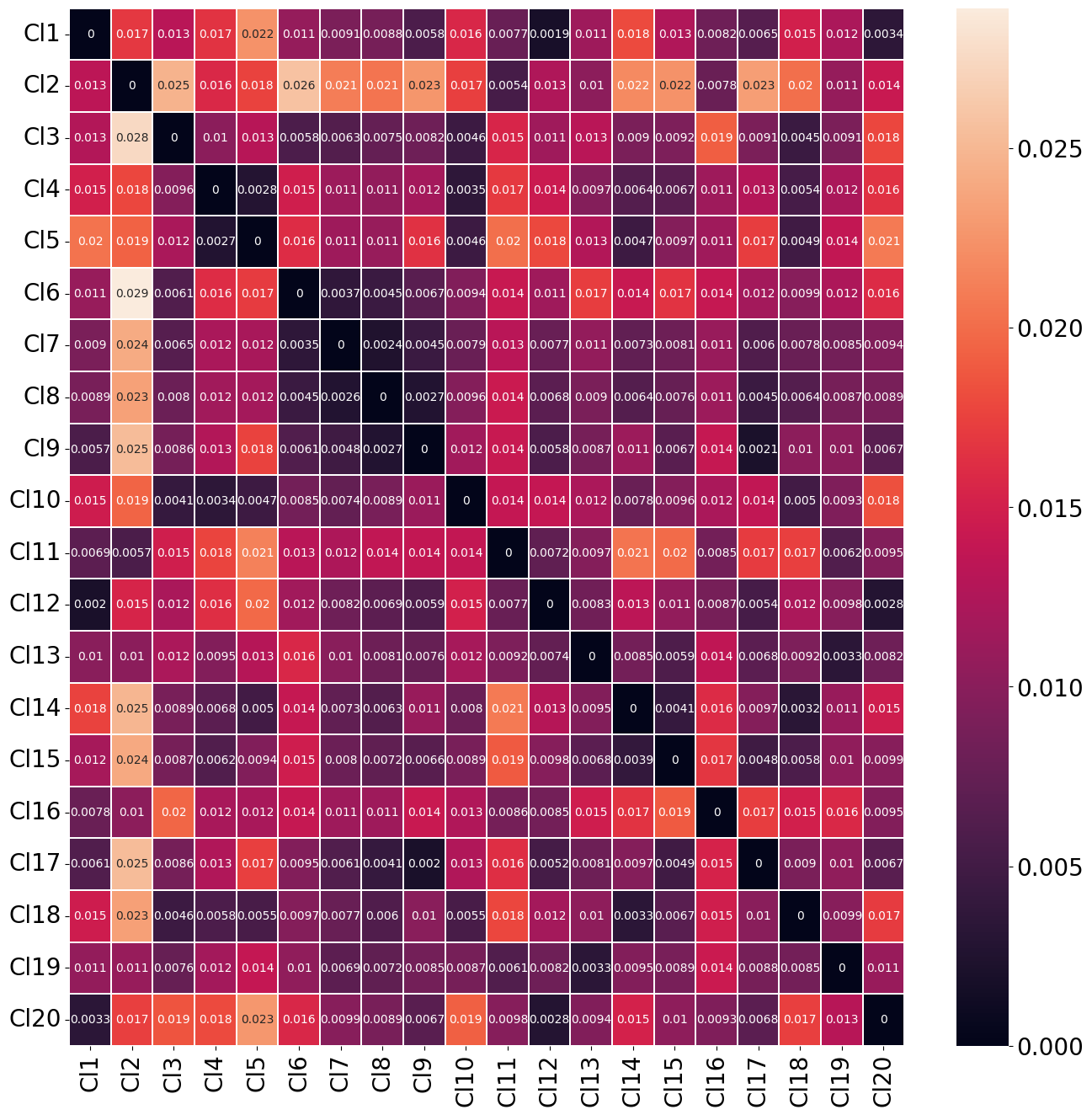} 
\caption{The similarity among 20 clients for Scenario-1.}
\label{mnist20}
\end{figure}

First, we address the traditional heterogeneity of sample distribution addressed by many of the previous FL works in Scenario-1.
We follow the same setup as mentioned in~\cite{byzantine} for this experiment. At the local client level, we train 20 cGAN models and share them with the cloud server. On the cloud server, we generate 1000 samples from each class per client to create an IID data distribution. Next, these datasets are used to calculate the nearest clients using the proposed AE-guided client similarity matrix. Fig.~\ref{mnist20} illustrates the calculated similarity matrix for Scenario-1. From the Figure, we see that the distance range is very narrow. The observation provides proof for the claim in the paper~\cite{rethinkheterogeneity}, that traditional data splits considered in previous works are almost IID and not challenging and contains a common feature space. This matrix is then used the derive $\tau_i$ for client $i$ and use the Eq. (~\ref{eq:7}) to generate required samples to create a synthetic dataset for the selected client from other clients. Then, the refined cGAN is trained on synthetic data and shared with each client. These GAN models are used to generate 50 samples per step, which are aggregated with the existing private data to train the classifier.

 The averaged best clients classification accuracy on the MNIST dataset is presented in Table~\ref{table:case1}. Our proposed Pl-FedGAN performs the best
 compared to the existing FL frameworks, which demonstrates the ability of our proposed method in traditional data heterogeneity and splits. On the other hand, it shows the validity of the proposed AE-guided client similarity measurement. Furthermore, aggregating clients on the server reduces the computational complexity and provides a personalized solution for different clients. 
Note that, since the client similarity matrix is asymmetric, each client neighbors determined by $\tau_i$ are unique.

\begin{figure}[t]
\centering
\includegraphics[width=0.4\textwidth, height=0.3\textwidth]{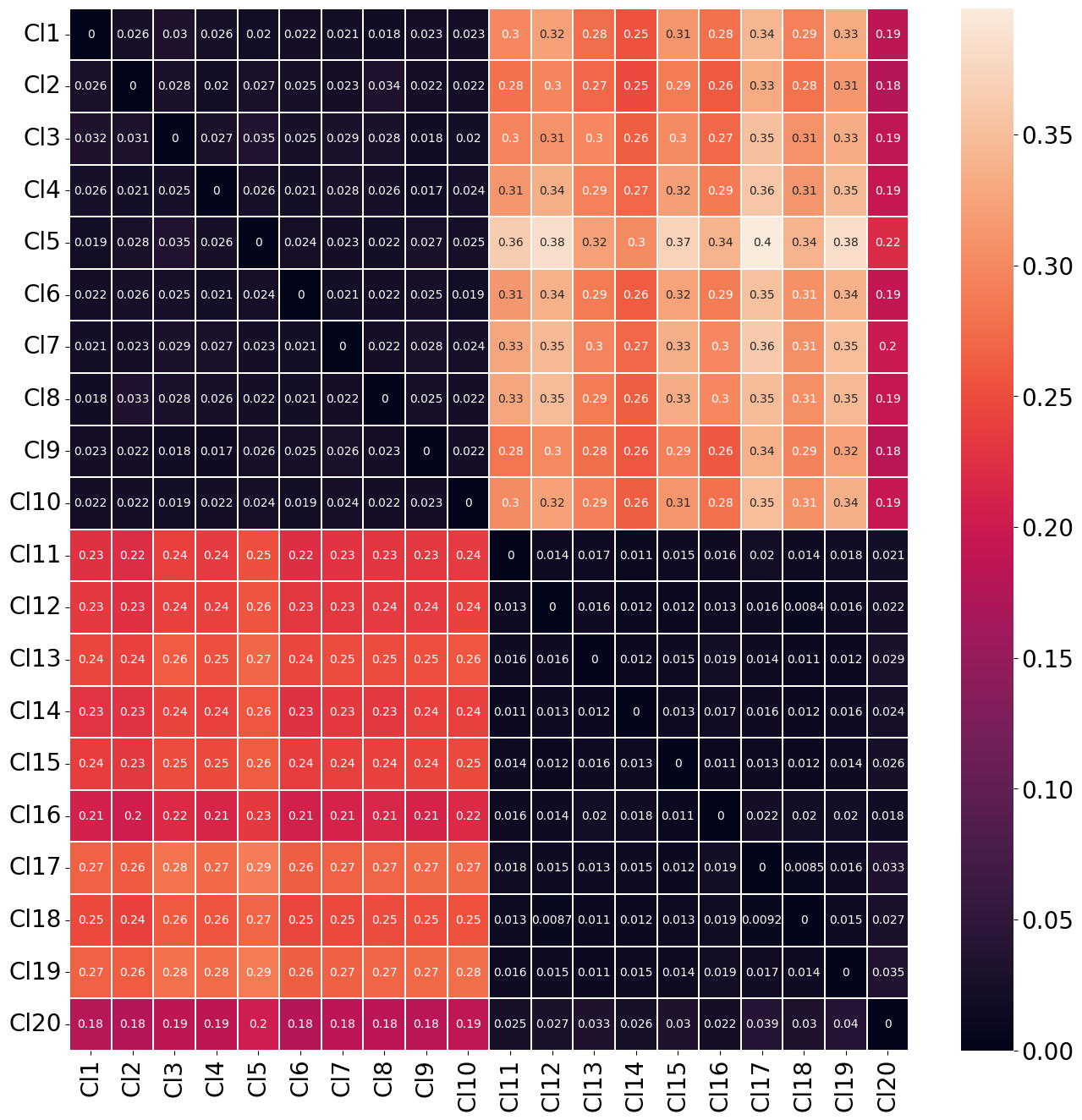} 
\caption{The similarity among 20 clients for Scenario-2. Here clients Cl1-Cl10 are from the MNIST dataset and the clients Cl11-Cl20 are from the FMNIST clients. The values clearly show the sensitivity of the proposed similarity measurement to different datasets. }
\label{mnist10fmnist10}
\end{figure}

\begin{table}[t]
\centering
\begin{tabular}{|l|l|}
\hline
Method & Accuracy \\
\hline
RFA-RFD  & 89.0 \\
Faug-FD & 88.0 \\
FedBN & 97.01\\
FedAvg & 91.06 \\
FedProx & 90.89\\
FedOptim & 91.93 \\
\hline
\textbf{PFL-GAN} & \textbf{97.11}\\
\hline
\end{tabular}
\caption{Classification accuracy on the downstream task for MNIST dataset. Some entries are from~\cite{byzantine}}
\label{table:case1}
\end{table}

\begin{table*}[t]
\centering
\begin{tabular}{|l|l|l|l|l|l|l|l|l|l|l|l|}
\hline
& Cl1 & Cl2& Cl3& Cl4& Cl5& Cl6& Cl7& Cl8& Cl9& Cl10& Mean\\
\hline

Cl1 & 
0.0&	0.0380&	0.0371&	0.0507&	0.0503&	\textbf{0.0946}&	\textbf{0.0711}&	\textbf{0.1013}&	\textbf{0.0919}&	\textbf{0.0915}&0.0696
 \\
Cl2 & 0.0404&	0.0&	0.0252&	0.0272&	0.0326&	\textbf{0.0963}&	\textbf{0.0856}&	\textbf{0.1054}&	\textbf{0.0914}&	\textbf{0.0990}&	0.0670
\\
Cl3 & 0.0389&	0.0236&	0.0&	0.0183&	0.0318&	\textbf{0.0637}&	\textbf{0.0510}&	\textbf{0.0634}&	\textbf{0.0550}&	\textbf{0.0601}&	0.0451
\\
Cl4 & 0.0522&	0.0259&	0.0188&	0.0&	0.0324&	\textbf{0.0940}&	\textbf{0.0658}&	\textbf{0.0820}&	\textbf{0.0729}&	\textbf{0.0830}&	0.0586
\\
Cl5 & 0.0544&	0.0323&	0.0328&	0.0347&	0.0&	\textbf{0.0959}&	\textbf{0.0777}&	\textbf{0.0862}&	\textbf{0.0843}&	\textbf{0.0963}&	0.0661
\\
Cl6 & \textbf{0.0835}&	\textbf{0.0808}&	\textbf{0.0556}&	\textbf{0.0768}&	\textbf{0.0802}&	0.0&	0.0211&	0.0331&	0.0251&	0.0162&	0.0525
\\
Cl7 &\textbf{0.0634}&	\textbf{0.0756}&	\textbf{0.0464}&	\textbf{0.0601}&	\textbf{0.0725}&	0.0201&	0.0&	0.0237&	0.0112&	0.0134&	0.0429
 \\
Cl8 &\textbf{0.0925}&	\textbf{0.0993}&	\textbf{0.0590}&	\textbf{0.0756}&	\textbf{0.0843}&	0.0357&	0.0242&	0.0&	0.0187&	0.0264&	0.0573
\\
Cl9 &\textbf{0.0805}&	\textbf{0.0821}&	\textbf{0.0485}&	\textbf{0.0644}&	\textbf{0.0807}&	0.0249&	0.0112&	0.0176&	0.0&	0.0141&	0.0471
\\
Cl10 & \textbf{0.0829}&	\textbf{0.0871}&	\textbf{0.0542}&	\textbf{0.0724}&	\textbf{0.0882}&	0.0153&	0.0130&	0.0261&	0.0146&	0.0&	0.0504
\\

\hline
\end{tabular}
\caption{The KLD-based similarity matrix for each client and mean distance. Bold values represent the distance greater than the mean distance of that client. Cli stands for the client-$i$. The first five clients represent the non-IID capital letter dataset whereas Cl6-Cl10 represents the clients with the non-IID simple letter dataset.}
\label{table:case3_kl}
\end{table*}

\subsection{Results of Scenario-2}
Next, we introduce the experimental results of Scenario-2.
Figure~\ref{mnist10fmnist10} displays the similarity matrix calculated for the Scenario-2 where clients Cl1-Cl10 contain the MNIST dataset, and clients Cl11-Cl20 are with the FMNIST dataset. From the results, we notice the range of the values is higher compared to Scenario-1. We observe a very close range of values among the clients from the same dataset, but a higher difference between the clients from different datasets, which display a clear boundary between the two different groups of clients. This result further validate the effectiveness of the proposed data similarity measurement.

To compare the classification accuracy, we evaluated each client with their corresponding test dataset, i.e., Cl1-Cl10 on the MNIST dataset and Cl11-Cl20 on the FMNIST dataset. 
The classification accuracies are presented in Table~\ref{table:case2}. From the results, we can see that in both cases our proposed Pl-FedGAN can outperform all the other existing methods by a large margin. The reason for such performance is that the proposed framework is capable of understanding the underlying client distribution and aggregating the clients from the same dataset. Among the same clients, the number of samples taken from each client varies due to the data heterogeneity between the clients representing the same dataset. From the similarity values, we see that data samples from different datasets form a challenging data heterogeneity problem limiting the performance of all the other existing methods. These results demonstrate the benefit of understanding the client underlying data distribution in collaborative learning via our proposed PFL-GAN and newly-defined client similarity.

\begin{table}[t]
\centering
\begin{tabular}{|l|l|l|}
\hline
Method & Acc. (MNIST) & Acc. (FMNIST) \\
\hline

FedBN & 75.46 & 67.13 \\
FedAvg & 75.80 & 30.28 \\
FedProx & 70.02 & 31.90\\
FedOptim & 81.96 & 69.38\\
\hline
\textbf{PFL-GAN} & \textbf{96.91} & \textbf{84.34}\\
\hline
\end{tabular}
\caption{Classification accuracy on the downstream task for MNIST dataset. In this experiment, we have used split-2 where we have 10 MNIST clients and 10 FMNIST clients. Two accuracy columns represent evaluating performances as the MNIST dataset on the baseline and the FMNIST dataset as the baseline respectively.}
\label{table:case2}
\end{table}

\begin{table}[t]
\centering
\begin{tabular}{|l|l|l|}
\hline
Method & Acc. (Capital) & Acc. (Simple) \\
\hline

FedBN & 95.70 & 89.03 \\
FedAvg & 95.32 & 89.075 \\
FedProx & 95.60 & 89.55\\
FedOptim & 95.42 & 88.52\\
Baseline* & 98.05 & 92.05\\
\hline
\textbf{PFL-GAN} & \textbf{97.58} & \textbf{91.03}\\
\hline
\end{tabular}
\caption{Classification accuracy on the downstream task for split-3. Baseline* represents the best classification accuracy obtained by training the classifier on the entire training set either in capital or simple letters and evaluated on the respective test set. Two accuracy columns represent evaluating performances on the capital letter dataset as the baseline and the simple letters dataset as the baseline respectively.}
\label{table:case3}
\end{table}

\begin{figure}[t]
\centering
\subfigure[Random samples from the capital letters dataset.]{\includegraphics[width=1.5in,height=2in]{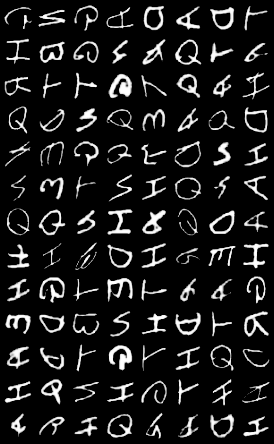}%
\label{fig3_1}}
\hfil
\subfigure[Random samples from the simple letters dataset. ]{\includegraphics[width=1.5in,height=2in]{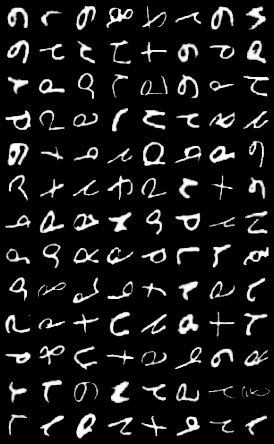}%
\label{fig3_2}}
\caption{Randomly selected samples from the Capital and Simple datasets.}
\label{fig_capsim}
\end{figure}
\begin{table*}[t]
\centering
\begin{tabular}{|l|l|l|l|l|l|l|l|l|l|l|l|}
\hline
& Cl1 & Cl2& Cl3& Cl4& Cl5& Cl6& Cl7& Cl8& Cl9\\
\hline

Cl1 & 
\textbf{0.0}&	\textbf{0.01023}&	\textbf{0.01141}&	0.17699&	0.24844&	0.21341&	0.16230&	0.16417&	0.14307

 \\
Cl2 & \textbf{0.01154}&	\textbf{0.0}&	\textbf{0.00544}&	0.17117	&0.23906&	0.20770	&0.15102&	0.15236&	0.13201

\\
Cl3 &\textbf{0.01227}&	\textbf{0.00522}&	\textbf{0.0}&	0.16685&	0.23642&	0.20491&	0.14910&	0.14949&	0.12832

\\
Cl4 & 0.30456&	0.28857&	0.27949&	\textbf{0.0}&	\textbf{0.02060}&	\textbf{0.01250}&	0.18363&	0.16642&	0.15163

\\
Cl5 & 0.43842&	0.42275&	0.41663&	\textbf{0.02167} &\textbf{0.0}&	\textbf{0.01034}	&0.23723&	0.22106&	0.21039

\\
Cl6 & 0.38459&	0.36722&	0.35898&	\textbf{0.01311}&	\textbf{0.01037}&
	\textbf{0.0}&	0.24137&	0.22295&	0.20571

\\
Cl7 &0.21845&	0.20279	&0.20579&	0.13883&	0.17009	&0.17476
&	\textbf{0.0}&	\textbf{0.00705}&	\textbf{0.01343}

 \\
Cl8 &0.21407&	0.19840&	0.19806	&0.13212&	0.16430	&0.16891&	\textbf{0.00697}
&	\textbf{0.0}&	\textbf{0.00549}

\\
Cl9 &0.18561&	0.17073	&0.16929&	0.12022&	0.15119&	0.15386&	\textbf{0.01396}&	\textbf{0.00540}
&	\textbf{0.0}
\\

\hline
\end{tabular}
\caption{The distance matrix for each client. Cl1-Cl3: MNIST dataset; Cl4-Cl6: FMNIST dataset; and Cl7-Cl9: Capital letter dataset. Bold values along a row represent the similar neighbors selected by K-Means algorithm.}
\label{table:ablation1}
\end{table*}

\subsection{Results of Scenario-3}
Next, we evaluate our performance in a Scenario-3, where clients are with the same label space but different feature space. In this setup, we select 10 clients, each client has access to all 10 classes from the simple letter dataset or capital letter dataset. We present the distance values measured between each client in Table~\ref{table:case3_kl}. Similar to the previous cases, we calculate $\tau$ by mean similarity to cluster the clients into 2 groups. In the same group, we see close values while we can identify a clear separation across the 2 groups. Another important observation is that the similarity value range is comparatively small. This results from the fact that these two datasets may have some shared feature space. Due to the transformations in the original dataset such as rotation, we can observe many similarities in the capital and simple letters leaving some samples ambiguous to categorize for the human eye. As shown in Fig.~\ref{fig_capsim}, some capital letters and their corresponding simple letters may have similar shapes such as `L' while some others display different textures.
Having a comparatively small range of similarity values, the above observation is highlighted by the similarity matrix as well.

We present the average best classification accuracy obtained by different FL methods in Table~\ref{table:case3}. We also report the best accuracy of the classifier trained on the entire training set ( therefore, IID without federation ) as baseline accuracy for comparison, which represents the highest possible accuracy the model can reach. From the table, we see that PFL-GAN performs the best compared to the other FL methods. PFL-GAN-FedGAN has reached almost the same accuracy as the baseline in both cases. Another interesting observation is that the difference between the baseline and other FL methods is not high as we see in the table~\ref{table:case2}. This observation can be explained using our similarity matrix, where the feature space of the two datasets has common features so that each client can benefit from each other.

\subsection{Ablation on the Number of Different Datasets} 
In previous experiments, we have considered clients from two different datasets to measure the worst-case performance of our proposed similarity measurement. In the following experiment, we formulate a PFL setup with 9 clients consisting of 3 clients representing each MNIST, FMNIST, and Capital letter dataset. All the clients have the aforementioned non-IID sample distribution. From Table~\ref{table:ablation1}, we can see that the similarity/distance measurement gives the smallest values for the clients with the same dataset. This experiment further provides proof for the expendability of our proposed distance measurement to cope with clients representing multiple datasets.

\subsection{Selection of $\tau$} \label{taus}
The selection of $\tau$ determines the grouping of neighborhood, which may influence the effectiveness of the proposed PFL-GAN. As a basic guideline, the choice of $\tau$ is governed by Eq (~\ref{eqn:tau}). 
To adpat our proposed PFL-GAN to the case of clients representing multiple datasets more than two, we propose an alternative to select $\tau$ and illustrate its performance here. 

Instead of defining an explicit $\tau$ to determine the client grouping and data aggregation, we cluster the distance matrix for each client $i$ into two groups and define $\tau_i^{'}$ as arbitrary value between the gap of two clusters. Then, we take all other client $j$ within the same cluster of $i$ for aggregation with sample number calculated by Eq. (\ref{eq:7}). For example, in Table~\ref{table:ablation1}, we apply K-Means clustering ~\cite{kmeans} for each row of the table to cluster clients into two groups. We bold the client's distance in the $i$th-row if it belongs to the same cluster of the client $i$. From the results, the new $\tau^{'}$ definition could group similar clients as expected.

\begin{figure}[h]
\centering
\includegraphics[width=0.3\textwidth, height=0.27\textwidth]{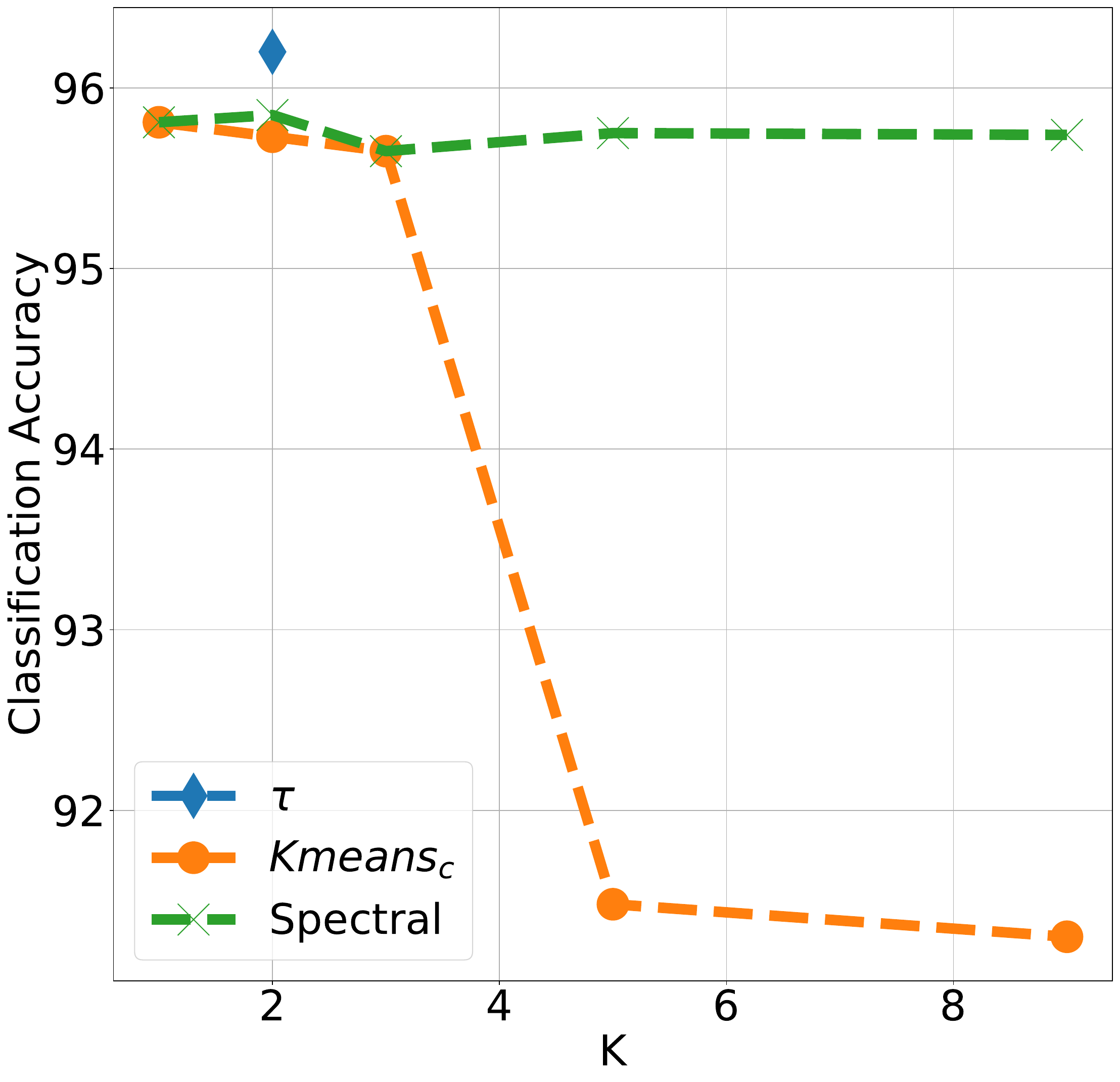} 
\caption{The classification accuracy over different numbers of clusters.}
\label{ablation}
\end{figure}

 We repeat the evaluation for Scenario-1,2,3 by applying K-Means to define $\tau_i^{'}$ for client grouping and data aggregation, which result in a similar clustering as $\tau$-based grouping in Eq. (\ref{eq:7}). Taking Scenario-1 as an example, the classification accuracy is $96.2\%$ with $\tau^{'}$, which is close to the accuracy obtained in Table~\ref{table:case1}. These results support the effectiveness of the proposed $\tau^{'}$. We also evaluate other alternative grouping strategies, such as the spectral clustering~\cite{spectral} and $K-means$ clustering to cluster clients into $K$ groups instead of a fixed $K=2$ in $\tau_i^{'}$. The classification results are shown in Fig.~\ref{ablation}, where $\tau$-based grouping provides the best performance. We plan to explore more efficient approaches for multi-dataset grouping in our future works.

\section{Conclusion}
In this work, we propose PFL-GAN, a more general framework for personalized learning to handle different types of data heterogeneity. More specially, we define a AE-guided similarity measurement with truncated RBF kernel to capture the underlying data distribution on the server from generated synthetic data. Our experimental results in demonstrate the effectiveness of PFL-GAN in addressing client data heterogeneity, including label skrewness, byzantine device detection and various feature spaces. In our future works, we plan investigate the estimation of multiple feature spaces among clients in FL and develop more efficient grouping strategy.



\bibliography{sample-bibliography}


\end{document}

%% file: dat620-setup.tex
\setcopyright{none}

\acmDOI{}

\acmISBN{}

\acmConference[Under review]{}
\copyrightyear{}
